\documentclass[conference]{IEEEtran}
\IEEEoverridecommandlockouts

\usepackage{cite}
\usepackage{amsmath,amssymb,amsfonts}
\usepackage{algorithmic}
\usepackage[ruled, linesnumbered]{algorithm2e}
\usepackage{graphicx}
\usepackage{textcomp}
\usepackage{xcolor}

\usepackage{amsthm}
\usepackage{multirow}
\usepackage{caption}
\usepackage{subfigure}
\usepackage[framemethod=tikz]{mdframed}
\definecolor{mygray}{RGB}{243,243,244}
\newmdenv[
  innertopmargin=0pt,
  backgroundcolor=mygray,
  linecolor=none,
  innerleftmargin=0pt,
  innerrightmargin=0pt,
  leftmargin=0pt
  ]{mymath}

\def\BibTeX{{\rm B\kern-.05em{\sc i\kern-.025em b}\kern-.08em
    T\kern-.1667em\lower.7ex\hbox{E}\kern-.125emX}}

\begin{document}

\title{DeepRicci: Self-supervised Graph Structure-Feature Co-Refinement for Alleviating Over-squashing}
\author{
\IEEEauthorblockN{Li Sun\IEEEauthorrefmark{1}, Zhenhao Huang\IEEEauthorrefmark{1}, Hua Wu\IEEEauthorrefmark{1}, Junda Ye\IEEEauthorrefmark{2}, Hao Peng\IEEEauthorrefmark{3}, Zhengtao Yu\IEEEauthorrefmark{4}, Philip S. Yu\IEEEauthorrefmark{5}}
\IEEEauthorblockA{
\IEEEauthorrefmark{1}\textit{School of Control and Computer Engineering, North China Electric Power University}, Beijing 102206, China \\
\IEEEauthorrefmark{2}\textit{School of Computer Science, Beijing University of Posts and Telecommunications}, Beijing 100876, China \\
\IEEEauthorrefmark{3}\textit{State Key Laboratory of Software Development Environment, Beihang University}, Beijing 100191, China \\
\IEEEauthorrefmark{4}\textit{Faculty of Information Engineering and Automation, Kunming University of Science and Technology}, Kunming 650500, China \\
\IEEEauthorrefmark{5}\textit{Department of Computer Science, University of Illinois at Chicago, IL, USA} \\
\{ccesunli, wuhua\}@ncepu.edu.cn, zhhuang@163.com, jundaye@bupt.edu.cn, penghao@buaa.edu.cn,\\
 ztyu@hotmail.com, psyu@uic.edu}
}

\maketitle


\begin{abstract}
Graph Neural Networks (GNNs) have shown great power for learning and mining on graphs, 
and Graph Structure Learning (GSL) plays an important role in boosting GNNs with a refined graph. 
In the literature, most GSL solutions 
either primarily focus on structure refinement with task-specific supervision (i.e., node classification),
or overlook the inherent weakness of GNNs themselves (e.g., over-squashing), resulting in suboptimal performance despite  sophisticated designs.
In light of these limitations, 
we propose to study self-supervised graph structure-feature co-refinement for effectively alleviating the issue of over-squashing in typical GNNs.
\emph{In this paper, we take a fundamentally different perspective of the Ricci curvature in Riemannian geometry}, in which we encounter the challenges of modeling, utilizing and computing Ricci curvature. 
To tackle these challenges, we present a self-supervised Riemannian model, \emph{DeepRicci}.
Specifically,
we introduce a latent Riemannian space of heterogeneous curvatures to model various Ricci curvatures, 
and propose a gyrovector feature mapping to utilize Ricci curvature for typical GNNs.
Thereafter, we refine node features by geometric contrastive learning among different geometric views, and simultaneously refine graph structure by backward Ricci flow based on a novel formulation of differentiable Ricci curvature.
Finally, extensive experiments on public datasets show the superiority of  DeepRicci, and the connection between backward Ricci flow and over-squashing.
Codes of our work are given in \emph{https://github.com/RiemanGraph/}.
\end{abstract}

\begin{IEEEkeywords}
Graph structure learning, Riemannian geometry, Ricci flow, Contrastive learning
\end{IEEEkeywords}


\section{Introduction}

Graphs are ubiquitous structures representing objects (nodes) and complex interactions among them, such as social networks, academic networks, biological networks, etc., 
and Graph Neural Networks (GNNs) is becoming the dominant solution for learning and mining on graphs \cite{DBLP:conf/iclr/KipfW17,DBLP:conf/nips/HamiltonYL17,DBLP:journals/tnse/LiuWDY22,DBLP:journals/tnn/ChenGWWXLLWL22}.
GNNs typically rely on one basic assumption that the input graph is reliable. 
However, this assumption is usually unrealistic. The graph structure is often noisy with missing or redundant edges \cite{DBLP:conf/kdd/LiL0CFY022,DBLP:journals/tnn/ChenGMWWWDLW22}.
Also, GNNs suffer from the side effect of message-passing paradigm  \cite{DBLP:conf/iclr/ToppingGC0B22}.
They both lead to the suboptimal performance of GNNs in the downstream tasks.
Thus, it becomes a hot research topic to learn an optimized graph for boosting GNNs, referred to as Graph Structure Learning (GSL).

While achieving encouraging results, it still presents several limitations.
On the one hand, both traditional supervised GSL solutions \cite{DBLP:conf/www/ZouPHYLWLY23,DBLP:conf/nips/WuZLWY22,DBLP:conf/kdd/SongZK22,DBLP:conf/www/LiuWWC0S22, DBLP:conf/cikm/SunLYFPJLY22} and the recent self-supervised ones \cite{DBLP:conf/www/LiuZZCPP22,DBLP:conf/wsdm/0002WJZ023} primarily focus on the structure refinement, i.e., refining the graph structure which tends to be noisy.
However, node features are often noisy as well.
This is problematic, because even if the structure is refined, GNN still learns from the noisy features, leading to suboptimal performance in downstream tasks.
Therefore, there is a need for the co-refinement of both graph structure and node feature to achieve better performance.
On the other hand, most GSL methods, including the recent \cite{DBLP:conf/www/LiuZZCPP22,DBLP:conf/wsdm/0002WJZ023}, overlook the inherent weakness of the message-passing within typical GNNs \cite{DBLP:conf/iclr/KipfW17,DBLP:conf/nips/HamiltonYL17,DBLP:conf/iclr/VelickovicCCRLB18}.
When a large volume of messages passes through a narrow path in the graph, the useful information tends to be distorted and squashed \cite{DBLP:conf/cikm/SunLYFPJLY22}.
This phenomenon, known as ``\emph{over-squashing}'', usually limits the expressive power of GNNs.
Besides, the supervised GSL \cite{DBLP:conf/cikm/SunLYFPJLY22} that relies on the supervision for node classification usually cannot benefit other applications, e.g., node clustering. On the contrary, self-supervised GSL is preferable for learning  generic knowledge.
In light of these limitations, 
we propose to study self-supervised graph structure-feature co-refinement for effectively alleviating the phenomenon of over-squashing in typical GNNs.

In Riemannian geometry, Ricci curvatures on a graph describe the high-order connectivity between node pairs \cite{RicciCurvatureOfGraph}, shedding lights on the structure learning.
Very recently, Ricci curvature is also connected to the phenomenon of over-squashing \cite{DBLP:conf/iclr/ToppingGC0B22,DBLP:conf/www/Liu0P00C023},
\textbf{motivating us to address GSL from a fundamentally different perspective of Riemannian geometry.}
However, we are facing the following challenges: 

\emph{Challenge 1: How to model Ricci curvature in the representation space?} 
Most representation spaces typically assume  a single uniform curvature radius, such as Euclidean, hyperbolic, spherical spaces and, more recently, quotient/ultrahyperbolic manifolds \cite{DBLP:conf/nips/XiongZPP0S22,DBLP:conf/nips/Law21}. 
However, Ricci curvatures indeed imply a representation space that exhibits  diverse curvatures over the edges/nodes.
In short, there is a lack of a promising space that enables different regions to have distinct curvatures, which is necessary for modeling Ricci curvature for GSL.

\emph{Challenge 2: How to utilize Ricci curvature for typical GNNs?} 
As mentioned above, Ricci curvature is built in Riemannian geometry, but the vast majority of GNNs \cite{DBLP:conf/iclr/KipfW17,DBLP:conf/nips/HamiltonYL17,DBLP:conf/iclr/VelickovicCCRLB18} work with the Euclidean space.
There is no isomorphism connecting the two types of spaces due to the inherent distinction in geometry \cite{RieGeo}. 
In other words, even if an appropriate representation space is designed, 
the challenge of bridging it to the Euclidean space for typical GNNs remains unresolved.

\emph{Challenge 3: How to compute Ricci curvature in a differentiable way?}
Ricci curvature is well defined in discrete problems, such as finding an optimal transport \cite{Olli2010Survey} or enumerating triangles \cite{DBLP:conf/iclr/ToppingGC0B22}.
However, solving discrete problem blocks the gradient backpropagation, preventing the joint learning of graph structure and curvature.
This poses the fundamental challenge to a learning model, and also explains why aforementioned methods \cite{DBLP:conf/iclr/ToppingGC0B22,DBLP:conf/www/Liu0P00C023} cannot be used for GSL.

\textbf{Present Work.}
To bridge the technical gaps, we propose a self-supervised GSL model, \textbf{DeepRicci}.
The novelty lies in that we introduce a \emph{latent Riemannian space}
in which we simultaneously refine graph structure and node features, alleviating the over-squashing on the graph.
To address challenge one, 
we construct the latent Riemannian space using a rotational product,  
and demonstrate that our design enables different nodes having different curvature, thereby facilitating the modeling of the various Ricci curvatures on the graph.
To address challenge two, 
we propose a gyrovector  feature mapping to bridge between the Riemannian manifold and Euclidean space with an isometry-invariant kernel.
To address challenge three, 
we introduce a differentiable formulation of Ricci curvature based on Olliver's definition \cite{Olli2010Survey}, replacing the discrete counterpart.
In reminiscent of the famous Ricci flow, \emph{graph structure is refined} in the backward Ricci flow with the differentiable curvature.
\emph{Node feature is simultaneously refined} in the contrasting learning among different geometric views induced from the product factors.
%
Furthermore, we prove that the graph learnt by DeepRicci 
\emph{alleviates the over-squashing by increasing the Cheeger’s constant on the graph}.

Overall, main contributions are summarized as follows.
\begin{itemize}
\item \emph{Problem.} In this paper, we propose a new problem of self-supervised graph structure-feature co-refinement, alleviating the over-squashing phenomenon in GNNs.
\item \emph{Methodology.} We take a fundamentally different perspective of Riemannian geometry, and propose the first Riemannian graph learning model (DeepRicci), to our best  knowledge. We introduce a latent Riemannian space, gyrovector  feature mapping and differential Ricci curvature with several theoretical guarantees. 
\item \emph{Experiment.} Extensive experiments on 4 public datasets show the superiority of DeepRicci.
 Furthermore, we empirically show the connection between Ricci flow and the over-squashing.
\end{itemize}

\begin{figure*}
\centering
    \includegraphics[width=0.95\linewidth]{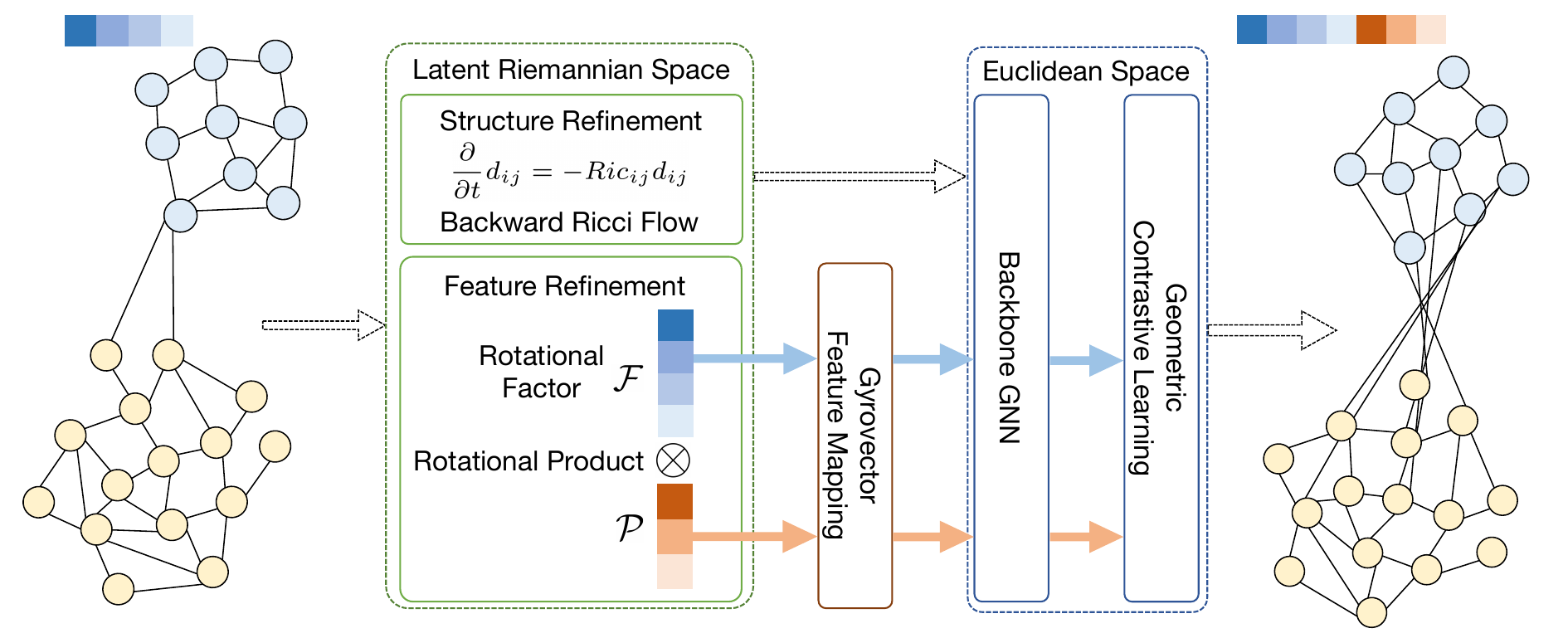}
            \vspace{-0.05in}
     \caption{Overall architecture of \textbf{DeepRicci}. We introduce a latent Riemannian space, constructed by rotational product (\textbf{Sec. \ref{TheSpace},B}), in which we jointly refine structure with backward Ricci flow (\textbf{Sec. \ref{DiffRiccCurv},D}) and refine feature by geometric contrastive learning (\textbf{Sec. \ref{SecOfFeatureRefine}}). A few linkages are added in the output graph  to alleviate the ``\emph{bottleneck}'' of the input graph (\textbf{Sec. \ref{theory}}).}
    \label{illu}
        \vspace{-0.2in}
\end{figure*}

       \vspace{-0.03in}
\section{Preliminaries}
        \vspace{-0.03in}
We first introduce the fundamental notion of manifold and curvature in Riemannian geometry.
Then, we put forward a new problem of \emph{self-supervised graph structure-feature co-refinement}, and point out its challenges.
        \vspace{-0.05in}
\subsection{Riemannian Geometry}
        \vspace{-0.03in}
\subsubsection{Manifold}
A smooth manifold $\mathcal M^d$ is said to be a Riemannian manifold if it is endowed with a Riemannian metric $g$, where $d$ is the dimensionality.
Each point $x \in \mathcal M$ is associated with a tangent space $\mathcal T_x\mathcal M^d \in \mathbb R^d$ around $x$ where the metric $g$ is defined, controlling the shape of the manifold.
Mapping between the tangent space and the manifold is done via exponential and logarithmic maps. 
In particular, the logarithmic map at $x$ does $log_x(\cdot): \mathcal M \to \mathcal T_x\mathcal M$, while the exponential map acts inversely.
\subsubsection{Curvature}
The curvature describes the extent how a surface deviates from being a plane. 
For $\boldsymbol x \in \mathcal M$, the sectional curvature $\kappa_x$ is defined on its tangent space $\mathcal T_x \mathcal M$.
A unit tangent vector $\boldsymbol v \in \mathcal T_x \mathcal M$ is associated with a Ricci curvature defined by averaging the sectional curvatures of planes containing $\boldsymbol v$, 
and the Ricci curvature controls the overlap of two balls w.r.t. the radii and the distance between their centers.



       \vspace{-0.03in}
\subsection{Problem Formulation}
       \vspace{-0.03in}
A graph $G(\boldsymbol A, \boldsymbol X)$ is defined on a node set $\mathcal V=\{v_1, \cdots, v_N\}$,
where $\boldsymbol A \in \mathbb R^{|\mathcal V|\times |\mathcal V|}$ is the binary adjacency matrix describing the graph structure. 
We have $[\boldsymbol A]_{ij}=1$ iff the nodes $v_i$ and $v_j$ are linked, otherwise, $[\boldsymbol A]_{ij}=0$.
Each node is attached with a feature $\boldsymbol x_i \in \mathbb R^F$, summarized in $\boldsymbol X \in \mathbb R^{|\mathcal V|\times F}$.
GNNs typically conduct message-passing over the graph structure with node features as the network input \cite{DBLP:conf/iclr/KipfW17,DBLP:conf/nips/HamiltonYL17,DBLP:conf/iclr/VelickovicCCRLB18}. 
GSL aims at boosting the performance of GNNs with an optimized graph. 
In this paper, we argue that \emph{not only graph structure but also node feature} affect the performance of GNN, given that both of them tend to be noisy in reality.
Thus, we explore a practical and challenging problem.
\newtheorem*{def1}{The Problem of Self-supervised Graph Structure-Feature Co-Refinement} 
\begin{def1}
Given a graph $G(\boldsymbol A, \boldsymbol X)$ with noisy structure $\boldsymbol A$ as well as noisy features $\boldsymbol X$,  
the target of self-supervised structure-feature co-refinement is to learn an optimized graph $G^\star(\boldsymbol A^\star, \boldsymbol X^\star)$ without supervision signals. 
The refined adjacency matrix $\boldsymbol A^\star \in [0, 1]^{N \times N}$ 
better encodes the underlying dependency between nodes with a latent Riemannian manifold $\mathcal M$, 
boosts the performance of downstream tasks  together with the refined node feature $\boldsymbol X^\star$, 
and  alleviates the over-squashing in the message-passing  of typical GNNs.
\end{def1}

In short, we are interested in jointly refining graph structure and node features, and alleviating over-squashing in absence of label information on the nodes or any downstream tasks.
\textbf{Fundamentally different from existing solutions, we approach GSL from a fresh perspective rooted in Riemannian geometry.}
Though Riemannian geometry provides elegant theory to study graph structures, e.g., Ricci curvature, we are still facing tremendous challenges. 

       \vspace{-0.03in}
\subsection{Challenges}
       \vspace{-0.03in}
\subsubsection{Homogeneous vs. Heterogeneous Space}
In Riemannian geometry, a smooth manifold is a \emph{homogeneous (curvature) space} if the sectional curvature is equal everywhere in the manifold.
A homogeneous space is characterized by the sectional curvature denoted as $\kappa$, since its value is regardless of $\boldsymbol x$. 
Correspondingly, the curvature radius is given as $\frac{1}{|\kappa|}$.
In particular, a $d-$dimensional homogeneous space is hyperbolic $\mathbb H^d_\kappa$ when $\kappa<0$. It is hyperspherical $\mathbb S^d_\kappa$ when  $\kappa>0$. 
Euclidean space $\mathbb R^d$ is a special case with $\kappa=0$. 
In the literature of graph data mining, most of the representation spaces are homogeneous  \cite{DBLP:conf/nips/ChamiYRL19,DBLP:conf/nips/LiuNK19,DBLP:conf/www/ZhangWSLS21,DBLP:conf/cvpr/DaiWGJ21,DBLP:conf/acl/ChenHLZLLSZ22}, including the recent studies \cite{DBLP:conf/nips/XiongZPP0S22,DBLP:conf/nips/Law21}.
They present a singe curvature radius $\kappa$, and thus lack the ability to model the various curvatures on the graph, posing \textbf{Challenge 1}.
Indeed, it calls for a \emph{heterogeneous (curvature) space} in which different points have different curvatures.
Furthermore, we are interested in boosting the performance of typical GNNs in the Euclidean space \cite{DBLP:conf/iclr/KipfW17,DBLP:conf/nips/HamiltonYL17,DBLP:conf/iclr/VelickovicCCRLB18}.
However, it still unclear how to bridge between a heterogeneous space and Euclidean space, and thus poses \textbf{Challenge 2}.

\subsubsection{Ricci Curvature}
The overlap of two balls is inherently connected to the cost of transportation moving one ball to the other, yielding the Ollivier definition of Ricci curvature \cite{Olli2010Survey}.
In a graph, 
given the mass distribution $m_u^\alpha(v)$  node $u \in \mathcal{V}$, 
\vspace{-0.03in}
\begin{equation}
    m_u^\alpha(v)=
    \begin{cases}
    \alpha, & v = u,\\
    (1 - \alpha)\frac{1}{|\mathcal N(u)|},& v \in \mathcal N(u), \\
    0, & Otherwise,
    \end{cases} \label{pmf}
    \vspace{-0.03in}
\end{equation}
where $\mathcal N(u)$ is the set of nodes linking to  $u$.
The Ollivier's Ricci curvature of a node pair $(i, j)$ is defined as follows:
\vspace{-0.03in}
\begin{equation}
    {Ric}^{\alpha}(i, j) = 1 - \frac{W(m_i^\alpha, m_j^\alpha)}{d(i, j)},
    \label{ORC}
    \vspace{-0.03in}
\end{equation}
where $W(m_i^\alpha, m_j^\alpha)$ is the Wasserstein distance \cite{DBLP:conf/nips/GulrajaniAADC17} between the probability (mass) distributions on nodes $i$ and $j$, and $d(i, j)$ is the distance defined on the graph.
The other popular variant is the Forman's Ricci curvature \cite{DBLP:conf/iclr/ToppingGC0B22}, whose definition involves triangle enumeration on the graph.
In other words, both definitions measure Ricci curvature with a discrete optimization problem, and thus blocks the gradient backpropagation for graph learning, posing \textbf{Challenge 3}.



\section{Our Approach: DeepRicci}

We present a self-supervised Riemannian model for graph structure-feature co-refinement, named as \textbf{DeepRicci}.
In a nutshell, we introduce a \emph{latent Riemannian space} with \emph{gyrovector  feature mapping},
in which we refine node features by the contrastive learning among different geometric views, 
and simultaneously refine graph structure with \emph{backward Ricci flow} based on a novel formulation of \emph{differentiable Ricci curvature}, alleviating the over-squashing on the graph.

The overall architecture of DeepRicci is sketched in Fig. \ref{illu}.

\noindent Before detailing the co-refinement mechanism, we first introduce the manifold  in which we build our model.

\vspace{-0.03in}
\subsection{Rotational Product Construction}\label{TheSpace}
\vspace{-0.03in}

We introduce a \emph{Latent Riemannian Space} $\mathcal M$, which is a heterogeneous manifold constructed by a carefully designed rotational  product,  addressing Challenge 1.

Specifically, we design the \emph{rotational  product} as follows,
\begin{equation}
\mathcal M = \mathcal P \otimes \mathcal F, \  \mathcal P=\otimes_{m=1}^{M-1}{\mathbb G}_{\kappa_m}^{d_m}, \ M \ge 2.
\end{equation}
$\mathcal M$ is the product of $M$ factor space, and the $M^{th}$ factor is the \emph{rotational factor}. $\otimes$ denotes the Cartesian product of manifolds. 
In the Cartesian product, a point in the product space $\boldsymbol x \in \mathcal M$ is expressed as 
\vspace{-0.03in}
\begin{equation}
\boldsymbol x=\operatorname{Concat}(\boldsymbol x^{\kappa_1}, \cdots, \boldsymbol x^{\kappa_m}, \cdots, \boldsymbol x^{\kappa_M}), \ \  \boldsymbol x^{\kappa_m} \in \mathbb G^{d_m}_{\kappa_m},
\vspace{-0.03in}
\end{equation}
where $\operatorname{Concat}(\cdot)$ denotes the operation of concatenation.
$\boldsymbol x^{\kappa_m} $ is the component of $\boldsymbol x$ in the $m^{th}$ factor.
The dimension of the product $\mathcal P$ is the sum of dimension of each factor $d=\sum\nolimits_{m=1}^{M-1} d_m$.
The distance between $\boldsymbol x$ and  $\boldsymbol y$ is given in  $d^2_{\mathcal P}(\boldsymbol x, \boldsymbol y)=\sum\nolimits_{m=1}^{M-1} d^2_{\kappa_m}(\boldsymbol x^{\kappa_m}, \boldsymbol y^{\kappa_m})$.
In the designed product, we utilize the \textbf{$\boldsymbol \kappa$-stereographic model} for each factor, \emph{unifying the vector operations of the aforementioned three types of spaces with gyrovector formalism.} 
Its $d$-dimensional model is a smooth manifold 
$\mathbb G^{d}_{\kappa}=\left\{\boldsymbol{x} \in \mathbb{R}^{d} \mid-\kappa |\boldsymbol{x}|^{2}<1\right\}$,
whose distance function is given as 
\begin{equation}
d(\boldsymbol x, \boldsymbol y)=\frac{2}{\sqrt{|\kappa|}}\tan^{-1}_\kappa\left(\sqrt{|\kappa|} |-\boldsymbol x \oplus_\kappa \boldsymbol y|\right),
\end{equation}
where $\oplus_\kappa$ is the gyrovector addition and $\tan^{-1}_\kappa$ is the curvature trigonometry \cite{DBLP:conf/icml/BachmannBG20}.
$\kappa$-stereographic model shifts to a hypersphere model of hyperspherical space with $\kappa > 0$, 
and to a Poincar{\'{e}} ball model of hyperbolic space with $\kappa<0$. 
In particular, the rotational factor is a upper hypersphere $\mathcal F=\bar{\mathbb G}_{+}^{d_M}$, where ${}_+$ denotes the positive curvatures.



\newtheorem*{pro1}{Proposition 1 (Heterogeneous Manifold)} 
\begin{pro1}
The product of $\mathcal M = \mathcal P \otimes  \mathcal F$ is a heterogeneous manifold, where $\mathcal F=\bar{\mathbb G}_{+}^{d_M}$ is the upper hypersphere providing curvature heterogeneity.
\end{pro1}

\begin{proof}
We sketch the proof with the key equations and omit math details owing to the limit of space.
\emph{$\mathcal M$ is heterogeneous iff the curvature $\kappa_{\boldsymbol x}$ is a function w.r.t. the location of $\boldsymbol x \in \mathcal M$}.
In the polar coordinates of $\bar{\mathbb G}_{+}^{d_M}$, Riemannian metric of rotational symmetry is $g_\rho = dr^2 + \rho^2(r)g_{\mathbb S^{d-1}}$ with some odd $\rho$ satisfying $\partial_r \rho(0)=1$. $r$ is the coordinate of the distance from the origin  (singular orbit) in gyrovector formalism.
The sectional curvature of $\bar{\mathbb G}_{+}^{d_M}$ is given by the function
\vspace{-0.05in}
\begin{equation}
\kappa_{\operatorname{rot}}=d_M \left(-2\frac{\partial^2_{rr}\rho}{\rho} + \left(d_M-1\right)\frac{1-(\partial^2_{r}\rho)^2}{\rho^2} \right), 
\vspace{-0.05in}
\end{equation}
according to \cite{RieGeo}(Chapter 3).
In the rotational product, the sectional curvature is expressed as 
\vspace{-0.05in}
\begin{equation}
\kappa_x=\kappa_{\mathcal P} + \kappa_{\operatorname{rot}}(r).
\vspace{-0.05in}
\end{equation}
The first term is derived as  a constant \cite{DBLP:journals/corr/abs-2202-01185}, and the second varies w.r.t. the location of $\boldsymbol x$, completing the proof.
\end{proof}

\noindent Given the polar coordinates of $\bar{\mathbb G}_{+}^{d_M}$, only $r-$coordinate contributes to curvature and distance $d_{\mathcal F}(\boldsymbol x, \boldsymbol y)=r_{\boldsymbol x}-r_{\boldsymbol y}$. 
For simplicity, we utilize the node feature $\boldsymbol x^{\operatorname{raw}}$ for $r-$coordinate, thus inducing $r=|\boldsymbol x|$ with  $d_{\mathcal F}(\boldsymbol x, \boldsymbol y)=|\boldsymbol x|-|\boldsymbol y|$.
\emph{It yields an equivalent product of $\mathcal M=\mathcal P \times \mathcal F$, where $\mathcal F \in \mathbb R^F$ is the rotational factor of Euclidean features, and $\mathcal P$ is the homogeneous product of hyperspherical or hyperbolic manifolds}. 


\noindent \textbf{Remarks.} It is noteworthy that the product construction itself \cite{DBLP:conf/iclr/GuSGR19,DBLP:conf/icml/BachmannBG20} cannot induce a manifold of heterogeneous curvatures. 
For instance, neither $\mathbb G^{d_1}_{-1} \times \mathbb G^{d_2}_{-2}$ (product of two hyperbolic manifolds) nor $\mathbb G^{d_1}_{-1} \times \mathbb G^{d_2}_{+1}$ (product of a hyperbolic manifold and hyperspherical one) will result in a constant curvature, as shown in the \emph{Proof}.
In other words, the product space without a rotational factor is still homogeneous.

\subsection{Gyrovector Feature Mapping}\label{LMap}

In this subsection, we build a bridge between Riemannian manifold and Euclidean space, addressing Challenge 2.
To this end, we propose a gyrovector  feature mapping $\phi^\kappa: \mathbb G_\kappa^{n} \to \mathbb R^{m}$, where $\mathbb G_\kappa^{n}$ is the Riemannian manifold of gyrovector formalism, and dimensions $n$ and $m$ can be different.

There is no isomorphism connecting the two types of spaces \cite{RieGeo}. 
Instead, we resolve this challenge with the aid of a kernel.
In the Euclidean space, for any isometry-invariant kernel, the eigenfunctions based on the Laplace operator construct (random) \emph{Fourier feature mapping} (the famous result of Bochner’s Theorem \cite{DBLP:conf/nips/RahimiR07}).
A popular parameterization for eigenfunctions are the \textbf{plane waves} $z_{\omega, b}(\boldsymbol x)=\sqrt{2}\cos(\langle \omega, \boldsymbol x\rangle +b)$, 
yielding a real-valued mapping with an invariant kernel $\mathbb E_{\omega, b}[z_{\omega, b}(\boldsymbol x)z_{\omega, b}(\boldsymbol y)]=k(\boldsymbol x-\boldsymbol y)$,
where $\omega$ and $b$ are uniformly sampled from a unit ball and $[0, 2\pi]$, respectively.


We generalize the aforementioned Fourier feature mapping to any Riemannian manifold.
The generalized feature mapping is constructed with \textbf{gyrovector waves}. For any  $\boldsymbol x^\kappa \in \mathbb G_\kappa^{n}$ of $\kappa-$stereographic model, the gyrovector wave is derived as
\vspace{-0.05in}
\begin{equation}
\operatorname{GF}^\kappa_{\lambda,b, \boldsymbol \omega}(\boldsymbol x^\kappa)=\exp\left(\frac{n-1}{2}\langle \boldsymbol \omega, \boldsymbol x^\kappa \rangle_\kappa\right)\cos\left( \lambda \langle \boldsymbol \omega, \boldsymbol x^\kappa \rangle_\kappa +b \right),
\label{gyroWave}
\end{equation}
where 
$
\langle \boldsymbol \omega, \boldsymbol x^\kappa \rangle_\kappa=\frac{1}{\sqrt{|\kappa|}} \tan^{-1}_\kappa\left(\sqrt{|\kappa|}\frac{|\boldsymbol x^\kappa|^2-\langle\boldsymbol \omega, \boldsymbol x^\kappa \rangle}{1+\langle\boldsymbol \omega, \boldsymbol x^\kappa \rangle} \right)
$
is indeed the signed distance.
We draw $m$ independent samples of  $\omega$, $b$  and $\lambda$ uniformly from the unit ball, $[0, 2\pi]$ and a distribution $\rho$,  respectively.
Gyrovector feature mapping is then given as  
\begin{equation}
\phi^\kappa(\boldsymbol x^\kappa)=\frac{1}{\sqrt{m}}\left[\operatorname{GF}^\kappa_{\lambda_1,b_1, \boldsymbol \omega_1}(\boldsymbol x^\kappa), \cdots,\operatorname{GF}^\kappa_{\lambda_m,b_m, \boldsymbol \omega_m}(\boldsymbol x^\kappa) \right],
\end{equation}
yielding the Euclidean $\phi^\kappa(\boldsymbol x^\kappa) \in \mathbb R^{m}$.


\newtheorem*{pro2}{Proposition 2 (Isometry Invariant)} 
\begin{pro2}
The induced kernel $k(\boldsymbol x^\kappa, \boldsymbol y^\kappa)=\mathbb E\left[ \operatorname{GF}^\kappa_{\lambda,b, \boldsymbol \omega}(\boldsymbol x^\kappa) \operatorname{GF}^\kappa_{\lambda,b, \boldsymbol \omega}(\boldsymbol y^\kappa)\right]$ from the gyrovector  feature mapping is isometry invariant and real.
\end{pro2}
\begin{mymath}
\begin{proof}
According to \cite{MR1790156}, the geometric identity for any isometry $g$ is given as
$
\langle g\circ \boldsymbol x^\kappa, g\circ \boldsymbol \omega \rangle=\langle  \boldsymbol x^\kappa, \boldsymbol \omega \rangle + \langle  g\circ \boldsymbol o, g\circ \boldsymbol \omega \rangle
$,
and thereby
$
\langle g^{-1}\circ \boldsymbol o, \boldsymbol \omega \rangle =- \langle  g\circ \boldsymbol o, g\circ \boldsymbol \omega \rangle
$.
By definition, 
\vspace{-0.05in}
\begin{equation}
k_\lambda(\boldsymbol x^\kappa, \boldsymbol y^\kappa)=\frac{1}{2}\mathbb E_{\boldsymbol \omega}\left[\exp\left(  n_\lambda^-  \langle \boldsymbol x^\kappa, \boldsymbol \omega \rangle_\kappa+n_\lambda^+\langle \boldsymbol y^\kappa, \boldsymbol \omega \rangle_\kappa\right)\right],
\end{equation}
where $n_\lambda^-=\frac{n-1}{2}-i\lambda$ and $n_\lambda^+=\frac{n-1}{2}+i\lambda$.
With $g\circ \boldsymbol y=\boldsymbol o$, 
\begin{equation}
\begin{aligned}
k_\lambda(g\circ \boldsymbol x^\kappa, \boldsymbol o)
& =  \frac{1}{2}\mathbb E_{\boldsymbol \omega}\left[\zeta_{\lambda,\boldsymbol \omega}(g\circ \boldsymbol x^\kappa) ^*\zeta_{\lambda,\boldsymbol \omega}(\boldsymbol o) \right]\\
& =\frac{1}{2}\mathbb E_{  \hat{\boldsymbol \omega}  }\left[\exp\left(  n_\lambda^-  \langle \boldsymbol x^\kappa, \hat{\boldsymbol \omega} \rangle_\kappa+n_\lambda^+\langle \boldsymbol y^\kappa, \hat{\boldsymbol \omega} \rangle_\kappa\right)\right],
\end{aligned}
\end{equation}
using $\hat{\boldsymbol \omega}=g^{-1} \circ \boldsymbol \omega$ as a change of variable,
where $\zeta_{\lambda,\boldsymbol \omega}(\boldsymbol z)$
$=\exp(n_\lambda^+ \langle \boldsymbol \omega, \boldsymbol z\rangle_\kappa)$ is the eigenfunction.
We have the equality
$k_\lambda(\boldsymbol x^\kappa, \boldsymbol y^\kappa)=k_\lambda(g\circ \boldsymbol x^\kappa, \boldsymbol o)$. 
For any $k_\lambda(\boldsymbol z, \boldsymbol o)$, it is derived as
\begin{equation}
k_\lambda(\boldsymbol z, \boldsymbol o)=\frac{1}{2}F_1\left(n_\lambda^+,n_\lambda^-;\frac{n}{2};\frac{1-\cos_\kappa(d_\kappa(\boldsymbol z, \boldsymbol o))}{2}\right),
\end{equation}
where $F$ is the hypergeometric function defined by power series \cite{Yu2022RandomLF}.
\emph{It claims that $k_\lambda(\boldsymbol x^\kappa, \boldsymbol y^\kappa)$ is distance-invariant depending on $d_\kappa(\boldsymbol z, \boldsymbol o)$, and thus isometry-invariant for any $\lambda$}.
Given a distribution $\rho$ over $\lambda$, the resulting kernel is 
\begin{equation}
\resizebox{1\hsize}{!}{$
k(\boldsymbol x^\kappa, \boldsymbol y^\kappa)=\frac{1}{2}\int_{-\infty}^{\infty} F_1\left(n_\lambda^+,n_\lambda^-;\frac{n}{2};\frac{1-\cos_\kappa(d_\kappa(\boldsymbol x^\kappa, \boldsymbol y^\kappa))}{2}\right)\rho(\lambda)d\lambda.
$}
\end{equation}
The kernel is shown to be \emph{real} with Legendre function.
\end{proof}
\end{mymath}

\noindent \textbf{Remarks.} Though the logarithmic map arrives at the tangent space $\mathcal T_x\mathcal M^d \in \mathbb R^d$, it is complicated and numerical unstable \cite{DBLP:conf/cvpr/DaiWGJ21}. 
No closed-form conclusions can be drawn without privileging of the origin (i.e., using the origin as the reference point for any point on the manifold) \cite{DBLP:conf/acl/ChenHLZLLSZ22}.
In contrast, our mapping is coupled with a isometry-invariant kernel, and we will show its empirical superiority in Ablation Study.
\cite{Yu2022RandomLF} constructs isometry-invariant mapping for hyperbolic space, but ours is suitable for both hyperbolic and hyperspherical space.

\vspace{-0.05in}
\subsection{Differentiable Ricci Curvature}\label{DiffRiccCurv}

The optimal transport problem nested in Ricci curvature blocks gradient backpropagation in graph learning (Challenge 3).
To bridge this gap, we formulate a novel differentiable Ricci curvature following the Ollivier's definition.

We first define a Laplacian matrix $\boldsymbol L^\alpha$ for structure $\boldsymbol A$ as
\vspace{-0.05in}
\begin{equation}
[\boldsymbol L^\alpha]_{ij}=
\begin{cases}
\alpha, & i=j,\\
(1-\alpha)\frac{1}{\boldsymbol D_{ii}}, & [\boldsymbol A]_{ij}=1, \\
0, & Otherwise,
\end{cases} \label{alphaLaplacian}
\vspace{-0.07in}
\end{equation}
whose matrix form is given as $\boldsymbol L^\alpha= \alpha \boldsymbol I + (1-\alpha) \boldsymbol D^{-1} \boldsymbol A$, where $\boldsymbol D$ is the diagonal degree matrix.
Then, we derive the following  Ricci matrix $\boldsymbol R^\alpha(\boldsymbol A, \boldsymbol X)$, whose $(i,j)-$element is the Ollivier's Ricci curvature of node pair $(i,j)$,
\vspace{-0.03in}
\begin{equation}
[\boldsymbol R^\alpha(\boldsymbol A, \boldsymbol X)]_{ij}=1- \frac{ [\boldsymbol L^\alpha f(\boldsymbol X)]_i-[\boldsymbol L^\alpha  f(\boldsymbol X)]_j }{d(\boldsymbol x_i, \boldsymbol x_j)},
 \label{Ric}
 \vspace{-0.03in}
\end{equation}
with an affine transform $f: \mathbb R^{N \times F} \to \mathbb R^N$. 
Note that, \emph{our formulation of Ricci curvature in Eq. (\ref{Ric})  is differentiable with respect to the node features $\boldsymbol x$.}

Next, we prove that our formulation is a well approximation of Ollivier's Ricci curvature in Eq. (\ref{ORC}).
\newtheorem*{pro3}{Proposition 3 (Upper Bound)} 
\begin{pro3}
The differentiable Ricci curvature in Eq. (\ref{Ric}) is the upper bound of Ollivier's Ricci curvature in Eq. (\ref{ORC}),  $[\boldsymbol R(\boldsymbol A, \boldsymbol X)]_{ij}=\sup Ric^{\alpha}_{ij}$.
\end{pro3}
\vspace{-0.07in}
\begin{mymath}
\begin{proof}
Applying Kantorovich-Rubinstein duality \cite{DBLP:conf/nips/GulrajaniAADC17}, the Wasserstein distance between two distributions is rewritten as 
\begin{equation}
    W(p, q) = \sup \nolimits_{\lVert f \rVert_L \leq 1} \mathbb{E}_{z\sim p}[f(z)] - \mathbb{E}_{z\sim q}[f(z)], \label{krduality}
\end{equation}
where $f$ is $1-$Lipschitz.
With Eq. (\ref{pmf}), Eq. (\ref{alphaLaplacian}) and Eq. (\ref{krduality}), 
\begin{equation}
    \begin{aligned}
        W(m_u^\alpha, m_v^\alpha) 
        &= \sup \limits_{\lVert f \rVert_L \leq 1} \sum_{w \in \mathcal{V}} f(w)m_u^\alpha(w) - \sum_{w \in \mathcal{V}} f(w)m_v^\alpha(w) \\
        &= \sup \limits_{\lVert f \rVert_L \leq 1} [\boldsymbol L^\alpha f(\boldsymbol X)]_i-[\boldsymbol L^\alpha  f(\boldsymbol X)]_j, 
    \end{aligned}
    \label{sup}
\end{equation}
It is easy to verify that affine transform is $1-$Lipschitz with proper scaling according to Cauchy-Schwartz inequality.
Thus, $f $ is feasible for the supremum problem. With Eq. (\ref{Ric}), Eq. (\ref{ORC}) and Eq. (\ref{sup}), we have $[\boldsymbol R(\boldsymbol A, \boldsymbol X)]_{ij}=\sup Ric^{\alpha}_{ij}$ hold.
\end{proof}
\end{mymath}

\vspace{-0.07in}
\noindent In particular, for the features $\boldsymbol X^m$ in the factor  $\mathbb G^{d_m}_{\kappa_m}$, the affine transform $f$ is parameterized as
\vspace{-0.05in}
\begin{equation}
f(\boldsymbol X^{\kappa_m})=\left(\boldsymbol W_m \otimes_{\kappa_m} \boldsymbol X^{\kappa_m}\right) \oplus_{\kappa_m} \boldsymbol b_m,
\vspace{-0.07in}
\end{equation}
where $\boldsymbol W_m$ and $\boldsymbol b_m$ denote the weight and bias. $\otimes_{\kappa_m}$ and $\oplus_{\kappa_m}$ are the gyrovector operations given in Appendix. Correspondingly, $f(\boldsymbol X)=\boldsymbol W_M \boldsymbol X+ \boldsymbol b_M$ for the Euclidean features.


\subsection{Structure Refinement with Ricci Flow}

In DeepRicci, we propose to refine graph structure in the latent Riemannian space in reminiscent of Ricci Flow.


In differential geometry, 
Ricci flow \cite{Hopper2010} evolves a smooth manifold so that regions of large positive curvature tend to be more densely packed than regions of negative curvature, and finally the manifold divides itself into several submanifolds. 
In analogy to the smooth manifold, a graph trends to narrow the bottleneck so as to finally divide itself into several packed subgraphs in the Ricci flow. 
\emph{Intuitively, the previous state graph backward Ricci flow has ``wider'' bottleneck and thus alleviates over-squashing.}
We provide theoretical analysis in Sec. \ref{theory}, and  visualize the graph evolvement of Ricci flow in the Case Study.

Structure refinement in our proposal is based on the intuition above. 
Specifically, the Ricci flow on a graph \cite{Olli2010Survey} is given as,
\begin{equation}
\frac{\partial }{\partial t} d(\boldsymbol x_i, \boldsymbol x_j)=-Ric_{ij} d(\boldsymbol x_i, \boldsymbol x_j),
\label{RicciFlow}
\end{equation}
where $d(\boldsymbol x_i, \boldsymbol x_j)$ is the distance in the manifold, and we use the differentiable formulation of Ricci curvature $Ric_{ij}$. 
In our design, the observed structure $\boldsymbol A$ is  inferred from \emph{the previous state of optimized structure} $\boldsymbol A^\star$ backward Ricci flow.
With Eqs.  (\ref{Ric}) and (\ref{RicciFlow}),  the distance in the observed graph is formulated,
\begin{equation}
\begin{aligned}
d(\boldsymbol x_i, \boldsymbol x_j) & =(1-[\boldsymbol R^{\alpha}(\boldsymbol A^\star, \boldsymbol X^\star)]_{ij})  d(\boldsymbol x^\star_i, \boldsymbol x^\star_j) \\
 & = [\boldsymbol L^\alpha(\boldsymbol A^\star) f(\boldsymbol X^\star)]_i-[\boldsymbol L^\alpha(\boldsymbol A^\star) f(\boldsymbol X^\star)]_j.
 \end{aligned}
 \label{backRicFlow}
\end{equation}
Note that, the distance in Eq. (\ref{backRicFlow}) is a function of  $\boldsymbol A^\star$ and $\boldsymbol X^\star$.
We learn the optimized structure $\boldsymbol A^\star$  by maximizing the likelihood of the observation as follows,
\begin{equation}
\mathcal L_{\operatorname{Struct}}=-\sum\nolimits_{i=1}^{N}\sum\nolimits_{j=1}^{N}\log p([\boldsymbol A]_{ij}=1| \boldsymbol A^\star).
\label{StructRefine}
\end{equation}
In particular, the likelihood is given by a Fermi-Dirac decoder \cite{DBLP:conf/nips/ChamiYRL19},
$ p([\boldsymbol A]_{ij}=1| \boldsymbol A^\star)=\operatorname{Sigmoid}\left(\frac{1}{s}\left( r-d(\boldsymbol x_i, \boldsymbol x_j)\right)\right)$, 
where $s$ and $r$ are parameters.

We consider the element in optimized structure as a random Bernoulli variable $[\boldsymbol A^\star]_{ij} \sim \operatorname{Ber}(\pi_{ij})$, parameterized by the pairwise similarity $\pi_{ij}$.
In latent Riemannian space, $\pi_{ij}$ takes the form of multihead cosine similarity as follows,
\begin{equation}
\pi_{ij}= \frac{1}{M}\sum\nolimits_{m=1}^{M}\cos\left(\boldsymbol S^m \phi^{\kappa_m}(\boldsymbol x^{\kappa_m}_i), \boldsymbol S^m \phi^{\kappa_m}(\boldsymbol x^{\kappa_m}_i) \right),
\end{equation}
where we have $M$ heads and $\boldsymbol S$ is the weight matrix. $\boldsymbol x^{\kappa_m}_i$ is the feature of Riemannian factor manifold, and $\phi^{\kappa_m}$  denotes the corresponding gyrovector  feature mapping. 
The issue is that $[\boldsymbol A^\star]_{ij}$ is not differentiable with respect to $\pi_{ij}$ in Bernoulli distribution.
Instead, we use the formulation below,
\begin{equation}
[\boldsymbol A^\star]_{ij}= \operatorname{Sigmoid}\left(\frac{1}{\tau}\left( \log \frac{\pi_{ij}}{1-\pi_{ij}}+\log \frac{\epsilon}{1-\epsilon}\right)\right),
\end{equation}
where $\epsilon$ is uniformly sampled from $(0,1)$ and $\tau \in \mathbb R^+$ for the Gumbel-sigmoid reparameterization \cite{DBLP:conf/sdm/Ng0FLC022}.
It gives the concrete relaxation of Bernoulli and, in this way, the binary $[\boldsymbol A^\star]_{ij}$ from a Bernoulli is transformed to a differentiable  function over $\epsilon$ and pairwise similarity $\pi_{ij}$.

\subsection{Feature Refinement with Geometric Contrastive Learning} \label{SecOfFeatureRefine}

In DeepRicci, we propose to simultaneously refine node feature by geometric contrastive learning, 
in which we consider the embedding generated from different Riemannian factors as different \textbf{geometric views}.

\emph{First, we model node features in the latent Riemannian space $\mathcal M$ with learnable parameters}.
Recall $\mathcal M=\mathcal P \otimes \mathcal F$, $\mathcal P = \otimes_{m=1}^{M-1} \mathbb G^{d_m}_{\kappa_m}$ where we utilize raw node feature $\boldsymbol x^{\operatorname{raw}}$ as the final factor $\mathcal F \in \mathbb R^F$ (Sec. \ref{TheSpace}).
For each node $v_i$, we have 
$\operatorname{Concat}(\boldsymbol x^{\kappa_1}, \cdots, \boldsymbol x^{\kappa_{M-1}}, \boldsymbol x^{\operatorname{raw}}) \in \mathcal M$, 
where $\boldsymbol x^{\kappa_m} \in \mathbb G^{d_m}_{\kappa_m}$ are learnable parameters, $m=1, \cdots, M-1$ and $M \ge 2$.

\emph{Second, we generate different geometric views from the product factors by a backbone GNN}.
The challenge here is that $\boldsymbol x^{\kappa_m}$ in the Riemannian factor cannot be fed into typical Euclidean GNNs.
Thanks to gyrovector  feature mapping  (Sec. \ref{LMap}), we map $\boldsymbol x^{\kappa_m}$ to Euclidean space, and generate  geometric view as follows,
\vspace{-0.05in}
\begin{equation}
\boldsymbol Z^{m} \gets \operatorname{GNN}\left(\boldsymbol A^\star, \phi^{\kappa_m}\left(\boldsymbol X^{\kappa_m}\right)| \Omega_{\operatorname{GNN}}\right),
\vspace{-0.05in}
\label{View}
\end{equation}
where $\Omega_{\operatorname{GNN}}$ is the parameter of backbone GNN, 
and $\boldsymbol A^\star$ is the optimized graph structure.
Node features in the factor manifold $\mathbb G^{d_m}_{\kappa_m}$ is summarized in $\boldsymbol X^{\kappa_m}$.
We utilize the gyrovector  feature mapping  $\phi^{\kappa_m}$, so that Euclidean GNN is able to generate the geometric view $\boldsymbol Z^{m}$.
In particular, we apply Fourier feature mapping  to generate the geometric view $\boldsymbol Z^{0}$.


\emph{Third, we learn the refined feature by contrasting among different geometric views.}
Concretely, we contrast each geometric view generated from Rimannian factor and the geometric view from Euclidean features, and vice versa.
The contrastive loss is given by the popular InfoNCE \cite{DBLP:journals/corr/abs-1807-03748} as follows,
\vspace{-0.03in}
\begin{equation}
\operatorname{MI}(\boldsymbol Z^0, \boldsymbol Z^{m})=-\frac{1}{N}\sum^{N}_{i=1} \log \frac{\exp( Sim( \boldsymbol z_i^0, \boldsymbol z_i^{m} \rangle)}{\sum^{N}_{j=1}\exp( Sim(\boldsymbol z_i^0, \boldsymbol z_j^{m}) )},
\label{fRefine1}
\end{equation}
where $N$ is the number of nodes. $Sim(\boldsymbol x, \boldsymbol y)$ denotes the similarity measure in Euclidean space, and we use the standard inner product for simplicity.
Thus, we have the loss of feature refinement as follows, 
\vspace{-0.03in}
\begin{equation}
\mathcal L_{\operatorname{Feature}} =\frac{1}{M-1}\sum\nolimits_{m=1}^{M-1} \left( \operatorname{MI}(\boldsymbol Z^0, \boldsymbol Z^{m}) +\operatorname{MI}(\boldsymbol Z^{m}, \boldsymbol Z^0) \right),
\label{fRefine2}
\end{equation}
where $M$ is the number of geometric views.



\textbf{Overall Objective.} Incorporating Eq. (\ref{StructRefine}) and Eq. (\ref{fRefine2}) , the overall objective of self-supervised DeepRicci is formulated as 
\vspace{-0.1in}
\begin{equation}
\mathcal L_{\operatorname{DeepRicci}}=\beta \mathcal L_{\operatorname{Struct}}+(1-\beta) \mathcal L_{\operatorname{Feature}},
\vspace{-0.03in}
\end{equation}
where $\beta$ is a balancing factor between structure refinement and feature refinement.
\textbf{\emph{Finally, we accomplish structure-feature co-refinement with the novel latent Riemannian space in a self-supervised fashion,  alleviating the over-squashing.}}

\textbf{Complexity Analysis.} The overall process of training DeepRicci is summarized in Algorithm 1.
We construct the latent Riemannian space in Line $1$, and Line $2$ bridges it to Euclidean space.
We refine structure in Line $5-6$ and refine feature in Line $8-9$.
The parameters of GSL and GNN are jointly learnt in Line $11$.
The space complexity of the algorithm is $O(N^2)$ owing to the multihead cosine similarity in Line $6$, while $N$ is the number of nodes.
The time complexity is yielded as $O(N^2)$ owing to the geometric contrastive learning in Line $9$.
In other words, the complexity order of  DeepRicci is the same as that of the recent self-supervised GSL methods \cite{DBLP:conf/www/LiuZZCPP22,DBLP:conf/wsdm/0002WJZ023}, while we alleviate the over-squashing for boosting the performance of backbone GNN.



\begin{algorithm}[t]
        \caption{     \textbf{Self-supervised DeepRicci  }      } 
        \KwIn{Graph $G(\boldsymbol A, \boldsymbol X)$, A Backbone GNN}
        \KwOut{Optimized $G^\star(\boldsymbol A^\star, \boldsymbol X^\star)$, Parameters  $\Omega_{\operatorname{GNN}}$
}
Initialize feature parameters $\boldsymbol x^{\kappa_m} \in \mathbb G^{d_m}_{\kappa_m}$;\\
\While{not converging}{   
        Perform feature map $\operatorname{GF}^\kappa_{\lambda,b, \boldsymbol \omega}(\cdot)$ for $\boldsymbol x^{\kappa_m}, \forall m$;\\
        \hfill $\rhd$ \emph{Structure Refinement}\\
        Compute differentiable Ricci curvature in Eq. (\ref{Ric});\\
        Refine structure with backward Ricci flow in Eqs. (\ref{backRicFlow})-(\ref{StructRefine}); \\
        \hfill $\rhd$ \emph{Feature Refinement}\\
        Generate geometric views with GNN in Eq. (\ref{View});\\
        Refine feature by contrasting among geometric  views in Eqs. (\ref{fRefine1})-(\ref{fRefine2}). ;\\
        \hfill $\rhd$ \emph{Jointly Learn $G^\star$ and $\Omega_{\operatorname{GNN}}$}\\
        Compute gradient $\nabla_{ \{\boldsymbol x^{\kappa_m}\}_m, \Omega_{\operatorname{GNN}} }\  \mathcal L_{\operatorname{DeepRicci}}$;\
}
\end{algorithm}

\begin{table*}
\vspace{-0.05in}
\caption{Node Classification on Cora, Citeseer, Chameleon and  Squirrel Datasets. (Best results: \textbf{Bold}, Runner up: \underline{\textit{Underlined}}.)}
\vspace{-0.05in}
\begin{tabular}{c  l | c c c | c c c | c c c | c c c }
\hline
& \multirow{2}{*}{\textbf{Method}  } & \multicolumn{3}{c|}{ \textbf{Cora}} & \multicolumn{3}{c|}{ \textbf{Citeseer}} & \multicolumn{3}{c|}{  \textbf{Chameleon}}& \multicolumn{3}{c}{  \textbf{Squirrel }} \\
&    & ACC & W-F1& M-F1  & ACC & W-F1& M-F1 & ACC & W-F1& M-F1 & ACC & W-F1& M-F1  \\
\hline
\multirow{9}{*}{\rotatebox{90}{    GCN } }   
& Original       
& 79.4\tiny{±0.4}  & 79.4\tiny{±0.4}  & 78.8\tiny{±0.3}  & 64.8\tiny{±0.6}  & 65.6\tiny{±0.6}  & 62.6\tiny{±0.6}  
& 32.2\tiny{±2.8}  & 30.5\tiny{±3.5}  & 30.5\tiny{±3.3}  & 22.2\tiny{±1.2}   & 21.9\tiny{±1.2}  & 21.9\tiny{±1.2} \\
& DropEdge         
& 80.4\tiny{±0.4}  & 80.8\tiny{±0.4}  & 79.4\tiny{±0.3}  & 67.6\tiny{±0.2}  & 68.3\tiny{±0.3}  & 65.0\tiny{±0.3}  
& 43.1\tiny{±0.5}  & 42.2\tiny{±0.8}  & 41.4\tiny{±0.9}  & 25.8\tiny{±1.2} & 25.1\tiny{±1.4} & 24.8\tiny{±1.5} \\
& IDGL               
& 78.7\tiny{±0.6}  & 78.7\tiny{±0.9}  & 77.7\tiny{±0.9}  & 66.0\tiny{±0.8}  & 66.8\tiny{±0.8}  & 63.2\tiny{±0.9}  
& 42.2\tiny{±1.8}  & 40.8\tiny{±1.7}  & 39.8\tiny{±1.7}  & 25.8\tiny{±2.2} & 13.8\tiny{±2.3} & 13.6\tiny{±2.2} \\
& PRI-GSL            
& 79.3\tiny{±0.4}  & 79.3\tiny{±0.4}  & 78.8\tiny{±0.6}  & 67.1\tiny{±1.0}  & 67.9\tiny{±0.9}  & 64.0\tiny{±0.9}  
& \underline{\textit{51.8}}\tiny{±0.2}  & \underline{\textit{51.3}}\tiny{±0.4}  & \underline{\textit{51.0}}\tiny{±0.5}  & 28.7\tiny{±1.2} & 27.7\tiny{±2.1} & 27.4\tiny{±2.2} \\
& PASTEL           
& \underline{\textit{82.6}}\tiny{±0.5}  & \underline{\textit{82.5}}\tiny{±0.3}  & 81.2\tiny{±0.3}  & \underline{\textit{72.1}}\tiny{±1.1} & \underline{\textit{72.3}}\tiny{±0.8}  & \textbf{69.3}\tiny{±0.9}  
& \textbf{57.8}\tiny{±1.6}  & \textbf{57.8}\tiny{±2.4}  & \textbf{57.3}\tiny{±2.4}  & \textbf{37.5}\tiny{±0.2} & \textbf{37.5}\tiny{±0.6} & \textbf{37.5}\tiny{±0.7} \\
\cline{2-14}
& SLAPS            
& 79.8\tiny{±1.0}  & 80.1\tiny{±0.7}  & 79.3\tiny{±0.2}  & 65.7\tiny{±0.5}  & 66.2\tiny{±0.5} & 63.8\tiny{±0.1}  
& 33.9\tiny{±1.3}  & 34.1\tiny{±0.2}  & 31.5\tiny{±1.8}  & 23.2\tiny{±0.7}  & 22.7\tiny{±0.3} & 21.6\tiny{±0.3} \\
& SUBLIME           
& 82.3\tiny{±0.4}  & \underline{\textit{82.5}}\tiny{±0.4}  & \underline{\textit{81.5}}\tiny{±0.4}  & 71.0\tiny{±1.0}  & 71.4\tiny{±1.0}  & 67.6\tiny{±1.0}  
& 38.7\tiny{±1.4}  & 37.6\tiny{±1.3}  & 37.4\tiny{±1.3}  & 25.6\tiny{±0.4}  & 23.8\tiny{±2.2} & 23.9\tiny{±2.1} \\
& GSR               
& 80.9\tiny{±0.2}  & 81.1\tiny{±0.2}  & 80.2\tiny{±0.02} & 69.0\tiny{±0.3}  & 68.8\tiny{±0.3}  & 65.1\tiny{±0.3}  
& 43.6\tiny{±0.8}  & 42.8\tiny{±0.9}  & 42.1\tiny{±0.9}  & 25.8\tiny{±0.4} & 23.5\tiny{±0.5} & 23.7\tiny{±0.5} \\
& \textbf{DeepRicci}
& \textbf{82.8}\tiny{±0.6} & \textbf{82.7}\tiny{±0.6}  & \textbf{81.9}\tiny{±0.4}  & \textbf{72.3}\tiny{±0.8}  & \textbf{72.5}\tiny{±0.7} & \underline{\textit{68.6}}\tiny{±0.4}  
& 48.6\tiny{±0.8}  & 48.6\tiny{±0.8}  & 48.2\tiny{±0.8}  & \underline{\textit{28.9}}\tiny{±0.6} & \underline{\textit{28.4}}\tiny{±0.7} & \underline{\textit{28.1}}\tiny{±0.7} \\
\hline
\multirow{9}{*}{\rotatebox{90}{    GAT} }     
& Original
& 77.9\tiny{±0.6}  & 78.3\tiny{±0.6}  & 77.7\tiny{±0.5}  & 63.9\tiny{±2.5}  & 64.4\tiny{±1.7}  & 60.6\tiny{±1.7}  
& 32.0\tiny{±2.6}  & 29.9\tiny{±3.5}  & 29.9\tiny{±3.1}  & 22.6\tiny{±1.2} & 20.5\tiny{±1.4} & 20.5\tiny{±1.4} \\
& DropEdge           
& 78.4\tiny{±0.6}  & 78.7\tiny{±1.3}  & 76.9\tiny{±1.5}  & 65.9\tiny{±0.4}  & 66.2\tiny{±0.4}  & 62.8\tiny{±0.5}  
& 32.1\tiny{±2.6}  & 30.3\tiny{±1.6}  & 30.2\tiny{±1.2}  & 26.0\tiny{±1.0} & 25.6\tiny{±1.1} & 25.6\tiny{±1.0} \\
& IDGL               
& 81.1\tiny{±1.0}  & 80.6\tiny{±1.0}  & 79.7\tiny{±0.9}  & 67.2\tiny{±1.3}  & 66.5\tiny{±1.5}  & 61.9\tiny{±1.8}  
& 49.1\tiny{±3.1}  & 48.4\tiny{±4.0}  & 47.8\tiny{±3.1}  & 29.5\tiny{±2.4} & 27.0\tiny{±2.6} & 27.0\tiny{±2.3} \\
& PRI-GSL            
& 79.7\tiny{±0.6}  & 79.7\tiny{±0.6}  & 78.5\tiny{±0.9}  & 66.1\tiny{±0.4}  & 66.2\tiny{±1.0}  & 62.6\tiny{±1.1}  
& 48.0\tiny{±2.4}  & 46.8\tiny{±2.3}  & 46.2\tiny{±2.4}  & 24.6\tiny{±0.9}  & 23.0\tiny{±2.1}  & 23.0\tiny{±2.9} \\
& PASTEL           
& 81.9\tiny{±1.4}  & \underline{\textit{81.9}}\tiny{±1.4}  & 80.7\tiny{±1.2}  & 66.8\tiny{±1.6} & 66.6\tiny{±1.9} & 62.0\tiny{±1.7}  
& \textbf{52.2}\tiny{±1.9}  & \textbf{52.1}\tiny{±2.7}  & \textbf{52.5}\tiny{±2.8}  & \underline{\textit{35.0}}\tiny{±1.1} & \underline{\textit{34.8}}\tiny{±1.3} & \underline{\textit{34.8}}\tiny{±1.3} \\
\cline{2-14}
& SLAPS            
& 78.2\tiny{±0.7}  & 78.3\tiny{±0.7}  & 77.3\tiny{±1.0}  & 64.8\tiny{±0.9} & 65.1\tiny{±1.2} & 61.7\tiny{±0.6}  
& 33.6\tiny{±1.8} &  29.5\tiny{±1.3}  & 28.4\tiny{±1.7}  & 19.0\tiny{±0.6} & 16.4\tiny{±0.3} & 15.8\tiny{±0.8}  \\
& SUBLIME            
&  \underline{\textit{82.2}}\tiny{±0.6} & 81.7\tiny{±0.7}  &  \underline{\textit{81.2}}\tiny{±0.6}   &  69.2\tiny{±1.2} &  \underline{\textit{69.7}}\tiny{±1.1}  &  \underline{\textit{66.0}}\tiny{±1.0}  
&  36.1\tiny{±0.6} &  35.6\tiny{±0.7}  &  35.6\tiny{±0.7}   &  24.7\tiny{±1.1} &  24.7\tiny{±1.1}  &  24.7\tiny{±1.1}  \\
& GSR               
& 81.8\tiny{±1.2}  & 81.8\tiny{±1.3}  & 80.9\tiny{±1.4}  & \underline{\textit{69.4}}\tiny{±1.0}  & 69.4\tiny{±0.9}  & 65.8\tiny{±0.9}  
& 37.8\tiny{±0.4}  & 36.0\tiny{±0.2}  & 35.7\tiny{±0.2}  & 26.7\tiny{±0.3} & 20.4\tiny{±0.1} & 20.5\tiny{±0.1} \\
& \textbf{DeepRicci}
& \textbf{82.2}\tiny{±0.5} &  \textbf{82.1}\tiny{±0.5}  & \textbf{81.6}\tiny{±0.2}  & \textbf{71.2}\tiny{±2.0} & \textbf{70.7}\tiny{±1.8} &  \textbf{68.9}\tiny{±0.5}       
& \underline{\textit{49.3}}\tiny{±1.3}  & \underline{\textit{49.1}}\tiny{±0.9}  & \underline{\textit{48.4}}\tiny{±1.1}  & \textbf{35.2}\tiny{±1.3} & \textbf{35.1}\tiny{±1.1} & \textbf{34.9}\tiny{±1.0} \\
\hline
\multirow{9}{*}{\rotatebox{90}{   SAGE} }
& Original                    
& 75.2\tiny{±1.3}  & 75.4\tiny{±1.6}  & 74.1\tiny{±1.6}  & 64.8\tiny{±1.8}  & 64.8\tiny{±1.6}  & 60.7\tiny{±1.6}  
& 39.9\tiny{±0.6}  & 39.2\tiny{±0.6}  & 39.5\tiny{±0.6}  & 27.6\tiny{±1.1} & 25.6\tiny{±1.5} & 25.6\tiny{±1.5} \\
& DropEdge          
& 79.9\tiny{±0.7}  & 80.5\tiny{±0.6}  & 79.1\tiny{±0.7}  & 66.7\tiny{±1.9}  & 67.7\tiny{±1.6}  & 63.9\tiny{±1.8}  
& 40.9\tiny{±1.1}  & 38.8\tiny{±1.0}  & 39.1\tiny{±0.9}  & 28.2\tiny{±1.0} & 26.1\tiny{±1.1} & 26.3\tiny{±1.0} \\
& IDGL               
& 79.6\tiny{±0.8}  & 79.2\tiny{±0.9}  & 78.4\tiny{±0.8}  & 65.3\tiny{±1.1}  & 65.6\tiny{±0.9} & 61.3\tiny{±1.2}  
& 41.6\tiny{±2.1}  & 41.2\tiny{±2.0}  & 41.0\tiny{±1.3}  & 35.1\tiny{±0.9} & 33.9\tiny{±0.9} & 33.7\tiny{±0.8} \\
& PRI-GSL          
& 79.9\tiny{±0.5}  & 80.0\tiny{±0.5}  & 79.1\tiny{±0.4}  & 65.6\tiny{±0.9}  & 65.9\tiny{±0.9}  & 62.2\tiny{±0.8}  
& 39.8\tiny{±0.3}  & 35.2\tiny{±1.5}  & 34.3\tiny{±1.4}  & 19.3\tiny{±0.5}  & 14.2\tiny{±1.5}  & 14.9\tiny{±1.4}  \\
& PASTEL           
& 81.0\tiny{±0.8}  & 81.1\tiny{±0.8}  & 79.8\tiny{±0.7}  & 65.7\tiny{±1.0} & 65.9\tiny{±1.1}  & 61.4\tiny{±1.4}  
& \textbf{47.5}\tiny{±0.7}  & \textbf{47.7}\tiny{±0.9}  & \textbf{46.9}\tiny{±0.9}  & \underline{\textit{35.6}}\tiny{±1.2} & \underline{\textit{34.3}}\tiny{±1.5} & \underline{\textit{34.3}}\tiny{±1.5} \\
\cline{2-14}
& SLAPS      
&  77.4\tiny{±1.1}  &  76.3\tiny{±0.8}    &  76.2\tiny{±0.8}     &  65.1\tiny{±2.8}  &  64.8\tiny{±0.7}    &  61.0\tiny{±0.9}
&  36.1\tiny{±2.2}  &  31.8\tiny{±1.2}    &  32.2\tiny{±1.0}     &  20.6\tiny{±1.5}  &  17.8\tiny{±0.6}    &  17.2\tiny{±0.8} \\      
& SUBLIME           
&  \underline{\textit{82.8}}\tiny{±0.5}  &  \underline{\textit{82.9}}\tiny{±0.4}  &  \underline{\textit{81.7}}\tiny{±0.4}   & \underline{\textit{70.4}}\tiny{±0.5} & \underline{\textit{70.8}}\tiny{±0.8}  & \underline{\textit{66.2}}\tiny{±1.2}  
&  36.6\tiny{±0.4}  &  35.3\tiny{±0.5}  &  35.5\tiny{±0.5}   & 28.0\tiny{±2.1} & 20.0\tiny{±1.4} & 19.9\tiny{±1.5} \\
& GSR                
& 81.0\tiny{±0.05} & 81.2\tiny{±0.06} & 80.3\tiny{±0.02} & 68.8\tiny{±0.05} & 68.7\tiny{±0.04} & 65.0\tiny{±0.04} 
& 39.1\tiny{±0.05} & 36.3\tiny{±0.06} & 35.8\tiny{±0.06} & 23.7\tiny{±0.2} & 21.5\tiny{±0.3} & 21.7\tiny{±0.3} \\
& \textbf{DeepRicci} 
&  \textbf{83.2}\tiny{±0.3} &  \textbf{83.3}\tiny{±0.7} &  \textbf{82.9}\tiny{±0.6} &  \textbf{72.0}\tiny{±0.1} &  \textbf{72.2}\tiny{±0.4} & \textbf{69.1}\tiny{±0.5}
&  \underline{\textit{42.2}}\tiny{±0.7} &  \underline{\textit{41.9}}\tiny{±0.7} & \underline{\textit{41.8}}\tiny{±0.8}  &  \textbf{35.9}\tiny{±1.2} &  \textbf{35.7}\tiny{±1.0} & \textbf{35.7}\tiny{±1.2} \\
\hline   
    \end{tabular}
    \label{classification}
    \vspace{-0.1in}
    \end{table*}

\section{Theory on Over-squashing}\label{theory}

We further elaborate on the theoretical aspects of the proposed DeepRicci, and show \emph{the theoretical guarantee on alleviating over-squashing on the graph}. 

Intuitively, useful information on the nodes is squashed when passing through the ``bottleneck'' on the graph, especially for the information from the long-distant nodes.
We employ the Cheeger's constant for  a formal discussion.
 \vspace{-0.03in}
\newtheorem*{def2}{Definition (Cheeger's Constant)} 
\begin{def2}
Given a $\mathcal G(\mathcal V, \mathcal E)$ with node set $\mathcal V$ and edge set $\mathcal E$, Cheeger's constant is defined as
 \vspace{-0.1in}
\begin{equation}
h_G= \min\nolimits_{\Omega \in \mathcal V}\frac{|\partial \Omega|}{\min\{vol(\Omega), vol(\bar{\Omega})\}},
 \vspace{-0.05in}
\end{equation}
where $\bar{\Omega}=\mathcal V \setminus \Omega$ is the complement of vertex set $\Omega$, and the boundary $\partial  \Omega=\{(x,y)\in\mathcal E: x\in \Omega, y\notin  \Omega \}$ is the set of all edges going from $\Omega$ to a node outside of $\Omega$.
\end{def2}
\noindent Cheeger's constant measures the degree of information squashing from nodes in $\Omega$ through its boundary, referred to as the ``bottleneck'' \cite{DBLP:conf/www/Liu0P00C023,DBLP:conf/iclr/ToppingGC0B22}. 
A smaller $h_G$ implies a narrowed bottleneck and severe over-squashing, while a larger $h_G$ alleviates over-squashing.
In fact, $h_G$ is controlled by the spectral gap.
\newtheorem*{lemma}{Lemma (Lower Bound \cite{Cheeger})} 
\begin{lemma}
The Cheeger's constant is lower bounded by $2h_G > \lambda_1$, where the spectral gap $\lambda_1$ is the first non-zero Laplacian eigenvalue of $G$.
\end{lemma}

In other words,  over-squashing is related to the spectral gap of the graph structure. We have the following proposition hold.


\newtheorem*{pro4}{Proposition 4 (Over-squashing)} 
\begin{pro4}
The  refined structure  of DeepRicci alleviates over-squashing by increasing the Cheeger’s constant on the graph.
\end{pro4}

\begin{mymath}
\begin{proof}
According to \cite{Olli2010Survey}, the inequality $\lambda_1 > \kappa_0$ holds for any finite graph, where $\kappa_0$ is the minimum positive Ricci curvature on the edges. That is, $2h_G > \lambda_1 > \kappa_0$, and $\kappa_0$ is increased in backward Ricci flow, which is proved in \cite{Hopper2010}.
\end{proof}
\end{mymath}
\vspace{-0.05in}
\noindent An intuitive check is that the bottleneck is narrowed in forward Ricci flow, as will be shown in Sec. \ref{Case} (Case Study).

\vspace{-0.05in}
\section{Results \& Discussion}
\vspace{-0.05in}
 In this section, we conduct extensive experiments to evaluate the proposed DeepRicci on $4$ public datasets with the aim of answering following research questions (\emph{RQs}):
\begin{itemize}
    \item \emph{\textbf{RQ1}}. How does DeepRicci boost the downstream tasks?
  \item \emph{\textbf{RQ2}}. How does each module contribute to DeepRicci?
  \item \emph{\textbf{RQ3}}.  How does Ricci flow connect to over-squashing?
\end{itemize}

\vspace{-0.05in}
\subsection{Experimental Setup}
\vspace{-0.03in}

\subsubsection{Datasets \& Baselines}
We choose the public datasets of Cora, Citeseer, Chameleon and Squirrel \cite{sun23priGSL,DBLP:conf/cikm/SunLYFPJLY22}.
\textbf{In this paper, we focus on the self-supervised GSL}.
Self-supervised baselines include the recent SUBLIME \cite{DBLP:conf/www/LiuZZCPP22}, and GSR  \cite{DBLP:conf/wsdm/0002WJZ023} of pretrain-finetune pipeline.
In addition, we include SLAPS  \cite{DBLP:conf/nips/FatemiAK21} which is trained by the denoising autoencoder only for self-supervised learning.
We also compare with \textbf{Supervised baselines}: 
DropEdge \cite{DBLP:conf/iclr/RongHXH20}, 
IDGL \cite{DBLP:conf/nips/0022WZ20},
the recent PRI-GSL \cite{sun23priGSL}, 
and PASTEL \cite{DBLP:conf/cikm/SunLYFPJLY22}  addressing over-squashing in supervised manner.
We do not compare with the graph rewriting methods \cite{DBLP:conf/iclr/ToppingGC0B22,DBLP:conf/www/Liu0P00C023}, as they cannot jointly learn with GNNs and thus different from our propose.
We evaluate each method with the popular backbones of GCN \cite{DBLP:conf/iclr/KipfW17}, GAT \cite{DBLP:conf/iclr/VelickovicCCRLB18} and SAGE \cite{DBLP:conf/nips/HamiltonYL17}.
\emph{Existing GSL methods are Euclidean, and we for the first time introduce Riemannian geometry to GSL, to the best of our knowledge.}

\vspace{-0.05in}
\subsubsection{Reproducibility}
We introduce the implementation details  to enhance reproducibility.
For simplicity, we instantiate DeepRicci with $3-$factor rotational product $\mathcal M=\mathcal P \otimes \mathcal F$, 
where $\mathcal P$ is the product of a hyperbolic $\mathbb G^{32}_{-1}$ and a hyperspherical  $\mathbb G^{32}_{+1}$, enjoying the merit of different geometries.
The coefficient $\beta$ is set to $0.5$ by default, $\rho$ is a Gaussian distribution in practice.
The parameters $\boldsymbol x^\kappa$ of Riemannian manifolds are learned by Riemannian Adam \cite{DBLP:conf/iclr/BecigneulG19}, while the others are learned by standard Adam optimizer.
We conduct experiment on the server of  Intel Xeon Gold 5120T CPU and NVIDIA Tesla V100 GPU.
We initialize $\boldsymbol x^\kappa$ randomly in the gyrovector balls of $\mathbb G^{32}_{-1}$ and $\mathbb G^{32}_{+1}$ (Line 1 in Algorithm 1).
Alternatively, it can be initialized as the embeddings learned from any Riemannian GNN, e.g., \cite{DBLP:conf/icml/BachmannBG20}.
We conduct $5$ independent runs for each model, and report the mean and standard derivations.
\textbf{Codes} and further implementation details are given in \emph{https://github.com/RiemanGraph/}.


\vspace{-0.1in}
\subsection{Main Results}
\vspace{-0.03in}
We evaluate our model on node classification and clustering.

\subsubsection{Node Classification}
Three popular metrics are employed: classification accuracy (ACC), Weighted-F1 score (W-F1)  and Macro-F1 score (M-F1) (\%).
We summarize the results on all the $4$ datasets in Table \ref{classification}.
``Original'' denotes the results of backbone GNN without GSL.
The supervised GSL methods optimize structure learning and node classification jointly, while self-supervised methods are employed in the same manner as \cite{DBLP:conf/www/LiuZZCPP22} to ensure a fair comparison.
In particular, GSR maintains its pretrain-finetune pipeline as described in the original paper \cite{DBLP:conf/wsdm/0002WJZ023}.
 \emph{\textbf{DeepRicci} consistently achieves the best results among self-supervised methods}, 
and even outperforms most supervised baselines, e.g., DeepRicci receives performance gain compared to the best supervised GSL on Cora.

\subsubsection{Node Clustering}
Three popular metrics are employed: clustering accuracy (ACC), Normalized Mutual Information (NMI) and Adjusted Rand Index (ARI) (\%).
Note that, the supervised GSL methods require node labels for learning an optimized graph, while clustering is unsupervised by nature.
Thus, supervised GSL methods cannot be applied to node clustering. 
In other words, only the self-supervised methods are suitable for node clustering, and we employ GCN as the backbone.
In particular, we utilize the pretrained embeddings of GSR \cite{DBLP:conf/wsdm/0002WJZ023} for clustering, as its finetune phase is designed for classification specifically.
We include K-Means as a reference, and also compare with a clustering method with the discrete Ricci curvature, RicciC \cite{RiccComm}.
In addition, we include recent \emph{Deep Graph Clustering} (DGC) methods:  CGC \cite{CGC}, gCooL\cite{gCooL}, HostPool \cite{HostPool} and AGC-DRR \cite{AGC-DRR}.
The results on Cora and Citeseer are reported in Table \ref{clustering}.
\emph{\textbf{DeepRicci} outperforms other GSL methods}, and consistently achieves the best results except NMI on Cora.
A reason is that the similarities in various geometric views of DeepRicci facilitate node clustering.


\begin{table}
\centering
\caption{Node Clustering (Best results: \textbf{Boldfaced}, Runner up: \underline{\textit{Underlined}}, Derivation is omitted for clarity.) }
\vspace{-0.05in}
\begin{tabular}{c l  | c c  c |c c c  }
\hline
&\multirow{2}{*}{\textbf{Method}  }  & \multicolumn{3}{c|}{ \textbf{Cora}} & \multicolumn{3}{c}{ \textbf{Citeseer}} \\
           &       & ACC & NMI & ARI  & ACC & NMI & ARI   \\
\hline
& K-Means &   $49.3$    & $32.5$    &$19.3$      & $54.4$      & $31.2$    &$28.5$   \\
\hline
\multirow{4}{*}{\rotatebox{90}{    DGC } } 
&CGC   & \underline{$\emph{72.7}$}& $56.1$    & $49.3$     &  $69.6$    &  $44.6$    & $45.0$  \\
&gCooL      &   $72.0$    & $55.3$    & $54.9 $     & \underline{$\emph{70.3}$}     & $45.8$  & \underline{$\emph{45.7}$} \\
&HostPool  &   $71.8$  & $\mathbf{60.2}$ &\underline{$\emph{55.3}$} & $70.1$  & \underline{$\emph{46.0}$}     & $45.2$  \\
&AGC-DRR&  $40.6$     & $19.3$   & $16.2 $     &  $68.3$    & $43.3$      & $44.1$ \\
\hline
& RicciC     &  $55.6$     & $54.8$    & $48.9 $      &  $67.3$    & $44.9$     & $44.2$ \\
\hline
\multirow{4}{*}{\rotatebox{90}{    GSL } } 
&SLAPS     &$ 69.2 $    & $54.9 $    &$37.4 $       & $67.1$    & $42.4$     & $39.2  $   \\
&Sublime   & $71.5$      & $54.1$    & $51.2$       &  $68.7$    & $44.3$     &  $43.9$  \\
&GSR         &$52.6 $      & $21.1$    & $24.0 $      & $59.2 $    & $29.5$     & $36.6 $   \\
&\textbf{DeepRicci}
 & $\mathbf{73.1}$ &\underline{$\emph{56.9}$}& $\mathbf{55.7}$ & $\mathbf{70.6}$  & $\mathbf{46.1}$ & $\mathbf{45.9}$ \\
\hline
\end{tabular} 
\vspace{-0.03in}
\label{clustering}
\end{table}

\begin{table}
\centering
\caption{Ablation Study on Classification.}
\vspace{-0.05in}
\begin{tabular}{c l | c c  c |c c c  }
\hline
& \multirow{2}{*}{\textbf{Variant}  } & \multicolumn{3}{c|}{ \textbf{Cora}} & \multicolumn{3}{c}{ \textbf{Citeseer}} \\
     &    & ACC & W-F1& M-F1  & ACC & W-F1& M-F1   \\
\hline
\multirow{4}{*}{\rotatebox{90}{GCN} } 
& \textbf{DeepRicci}
& $\mathbf{82.8}$ & $\mathbf{82.7}$ & $\mathbf{81.9}$ & $\mathbf{72.3}$  & $\mathbf{72.5}$ & $\mathbf{68.6}$  \\
& $-Gyro$  &$80.4 $&$81.0$ & $80.1$&$71.2$ & $72.0$&$ 67.9$  \\
& $-Ricci$   &\underline{$\emph{81.9}$} & \underline{$\emph{82.1}$} &\underline{$\emph{81.1}$} &$71.5$ &\underline{$\emph{72.2}$} & $68.3$  \\
& $-Feature$  &$80.8$ &$81.0$ & $80.3$ &\underline{$\emph{71.7}$} &$72.1$ & \underline{$\emph{68.6}$} \\
\hline
\multirow{4}{*}{\rotatebox{90}{SAGE} } 
& \textbf{DeepRicci}
    &  $\mathbf{83.2}$ &  $\mathbf{83.3}$ & $\mathbf{82.9}$ &  $\mathbf{72.0}$ & $\mathbf{72.2}$ &  $\mathbf{69.1}$ \\
& $-Gyro$ &$81.9 $ & $82.0 $& $ 80.3$&\underline{$\emph{70.1}$} & \underline{$\emph{70.4}$}&  \underline{$\emph{67.2}$} \\
& $-Ricci$     & \underline{$\emph{83.1}$} & $82.8 $& \underline{$\emph{82.6}$}&$69.3 $ & $69.7 $&  $66.6 $ \\
& $-Feature$ & $82.7 $ & \underline{$\emph{83.1}$}& $ 81.5$&$69.6 $ & $70.0 $&  $66.9 $ \\
\hline
\end{tabular} 
\label{ablation-classification}
\vspace{-0.2in}
\end{table}

\vspace{-0.07in}
\subsection{Ablation Study}
\vspace{-0.05in}
We examine the effectiveness of proposed components of DeepRicci: \emph{gyrovector  feature mapping}, \emph{backward Ricci flow} and \emph{feature refinement}.
To evaluate the gyrovector features, we introduce a variant (denoted as $-Gyro$) that replaces the gyrovector features $\phi^\kappa(\boldsymbol x^\kappa)$ with the logarithmic mapped features $\log^\kappa_{\boldsymbol 0}(\boldsymbol x^\kappa)$. 
The expression of $\log^\kappa_{\boldsymbol 0}(\cdot)$ is given in the Appendix.
The reference point of logarithmic map is the origin $\boldsymbol 0$, and note that  logarithmic map cannot take the point $\boldsymbol x^\kappa$ as the reference point \cite{DBLP:conf/icml/BachmannBG20}.
To evaluate Ricci flow, we introduce a variant (denoted as $-Ricci$) that disables the backward Ricci flow in structure refinement by setting $\beta=0$. 
To evaluate the significance of feature refinement, we introduce a variant (denoted as $-Feature$) that does not concatenate the gyrovector features $\phi^\kappa(\boldsymbol x^\kappa)$ for downstream tasks.
We conduct ablation study on classification with the backbones of GCN and SAGE, and on clustering with GCN backbone.
The results on Cora and Citeseer are reported in Table \ref{ablation-classification} (classification) and Table \ref{ablation-clustering} (clustering), and we find that 
\textbf{(1)} The gyrovector features outperform the logarithmic features in all the cases.
Gyrovector features bridge Riemannian manifold and Euclidean space with an isometry-invariant kernel (\emph{Proposition 2}), while logarithmic map cannot generate Euclidean vectors with such guarantee.
\textbf{(2)} The model with backward Ricci flow achieves better results. We will further study on this point in the Case Study.
\textbf{(3)} Feature refinement generally boosts the performance of GNNs.
This is reasonable, as GNNs pass the message on the nodes, and noisy node feature trends to yield supoptimal embeddings for downstream tasks.
\emph{It verifies the motivation of our study that refining the noisy features further boosts the performance of backbone GNN.}

\begin{table}[t]
\centering
\caption{Ablation Study on Clustering.}
\vspace{-0.05in}
\begin{tabular}{c l  | c c  c |c c c  }
\hline
&\multirow{2}{*}{\textbf{Variant}  }
                       & \multicolumn{3}{c|}{ \textbf{Cora}} & \multicolumn{3}{c}{ \textbf{Citeseer}} \\
 &                  & ACC & NMI & ARI  & ACC & NMI & ARI   \\
\hline
\multirow{4}{*}{\rotatebox{90}{    GCN } } 
& \textbf{\textbf{DeepRicci}}
                       & $\mathbf{73.1}$ & $\mathbf{56.9}$& $\mathbf{55.7}$ & $\mathbf{70.6}$  & $\mathbf{46.1}$ & $\mathbf{45.9}$ \\
& $-Gyro$& $70.9$  &  \underline{$\emph{55.2}$} &  $49.6$  &  \underline{$\emph{69.3}$} &  $43.9$  &  $44.7$  \\
& $-Ricci$  & \underline{$\emph{72.7}$} & $55.0$ &  \underline{$\emph{52.1}$}  &  $69.1$ &  \underline{$\emph{44.6}$}  &  \underline{$\emph{45.3}$}  \\
& $-Feature$ & $71.3$  &  $54.8 $ &  $49.2$  &  $68.7$ &  $44.2$  &  $45.1$  \\
\hline
\end{tabular} 
\label{ablation-clustering}
\end{table}


\begin{figure} 
 \vspace{-0.15in}
\centering 
\subfigure[Backward Ricci Flow]{
\includegraphics[width=0.75\linewidth]{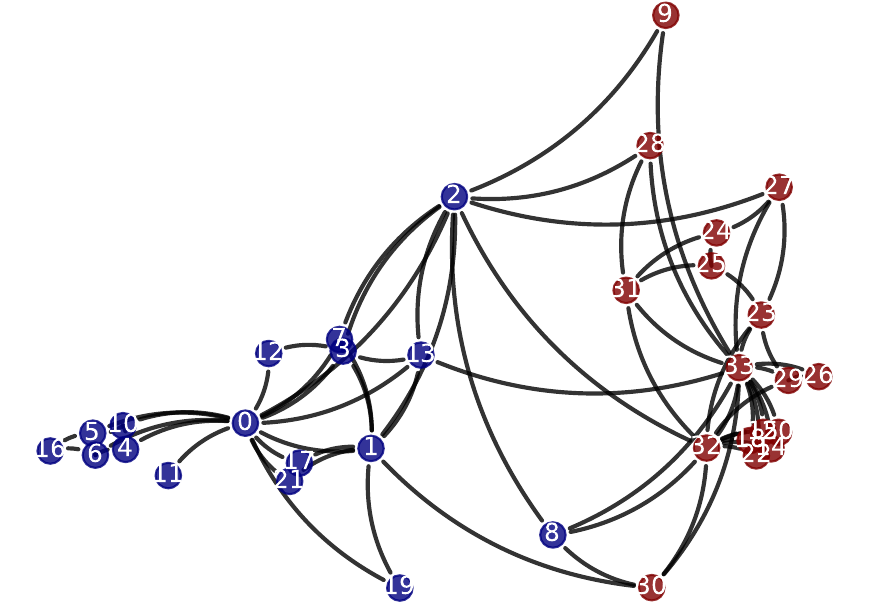}  \vspace{-0.2in}}
\subfigure[Forward Ricci Flow]{
\includegraphics[width=0.75\linewidth]{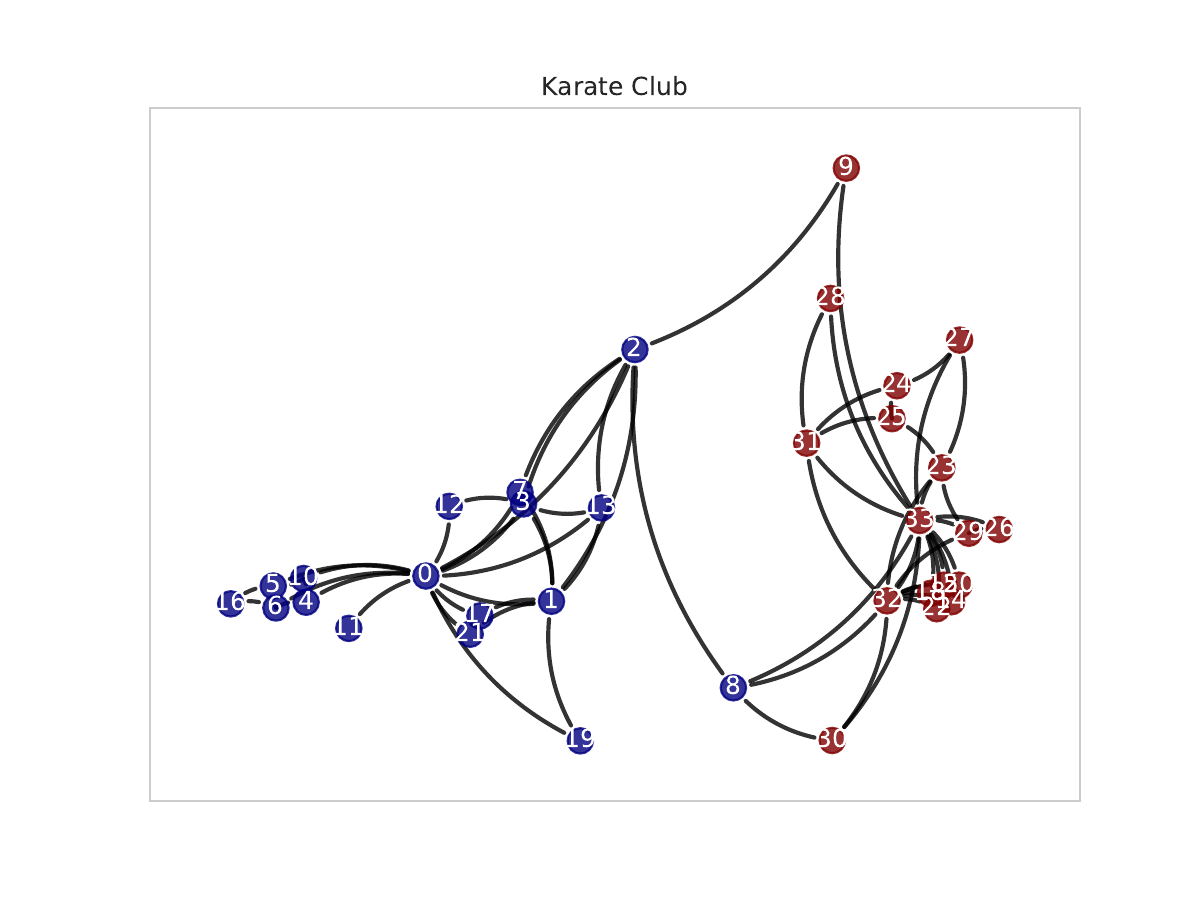}  \vspace{-0.1in}}
 \vspace{-0.1in}
\caption{Visualization of Karate Club Network.}
 \vspace{-0.25in}
\label{CaseStudy}
\end{figure}

\vspace{-0.1in}
\subsection{Case Study: Ricci Flow \& Over-squashing} \label{Case}
\vspace{-0.05in}
Here, we empirically study the connection between Ricci flow and over-squashing.
To this end, we conduct case study on Karate Club network, visualized in Fig. \ref{CaseStudy}.
The backward subfigure is graph structure learnt by DeepRicci, while the forward subfigure is generated by another variant model where backward Ricci flow is replaced by a forward one.
Karate Club network has two communities, distinguished by the different colors in Fig. \ref{CaseStudy}.
The network evolves itself through forward Ricci flow to a state where two communities link to each other by a few edges. 
Consequently, information in the community is squashed when passing through such a narrow bottleneck.
\emph{Under backward Ricci flow, the bottleneck is ``widened'' by more linkages, motivating our design for structure refinement.}
Indeed, we further prove that backward Ricci flow alleviates over-squashing in Sec. \ref{theory}.

\vspace{-0.07in}
\section{Related Work}
\vspace{-0.05in}
\subsection{Graph Structure Learning}
\vspace{-0.05in}
In the literature, learning graph structure has been studied in traditional machine learning, e.g., spectral clustering and graph signal processing \cite{DBLP:conf/wsdm/WangZF22}.
In this paper, we focus on the Graph Structure Learning (GSL) \emph{for boosting GNNs with an optimized graph}.
One line of work is driven by pairwise similarity.
The similarity is either measured by a kernel \cite{DBLP:conf/nips/0022WZ20}, 
or parameterized with a neural model \cite{sun23priGSL}.
\cite{DBLP:conf/nips/FatemiAK21} further improves GSL with a denoising autoencoder.
Gumble reparameterization trick is frequently used for optimizing the discrete graph structure \cite{DBLP:conf/nips/WuZLWY22}. 
Another line of work directly parameterizes the adjacency matrix, 
and training methods include bilevel optimization \cite{DBLP:conf/ijcai/HuCMS22}, 
Bayesian approach \cite{DBLP:conf/www/WangM0XJS021},
and variational inference \cite{DBLP:conf/kdd/SongZK22}.
Recent studies \cite{DBLP:conf/www/LiuZZCPP22,DBLP:conf/www/LiuWWC0S22,DBLP:conf/wsdm/0002WJZ023} incorporate contrastive learning  in GSL.
\cite{DBLP:conf/icml/LiuCSJ22} introduces an edge-to-vertex transform and its inverse.
\cite{DBLP:conf/www/ZouPHYLWLY23} conducts GSL by structural entropy.
\cite{DBLP:conf/icml/XieXJ22} infers a latent graph for self-supervised representation learning.
In addition, researchers study GSL on bipartite graphs \cite{DBLP:conf/nips/WeiLLW22}, 
heterogeneous graphs \cite{DBLP:conf/aaai/ZhaoWSHSY21}, 
hypergraphs \cite{DBLP:conf/ijcai/CaiSSZH022}, etc.
We specify that the recent \cite{DBLP:conf/cikm/SunLYFPJLY22} alleviates over-squashing in Euclidean space, and is supervised by node labels.
To the best of our knowledge, we for the first time address the self-supervised graph structure-feature co-refinement, alleviating over-squashing with theoretical guarantees.

\vspace{-0.07in}
\subsection{Riemannian Geometry \& Graphs}
\vspace{-0.03in}
Riemannian geometry has emerged as a powerful tool to model graphs.
Among the Riemannian spaces, hyperbolic space is well aligned with the graphs with hierarchical or tree-like structures \cite{DBLP:conf/nips/LiuNK19,DBLP:conf/icml/00010Y0K23}, 
and hyperbolic GCNs of different formulations are introduced \cite{DBLP:conf/nips/ChamiYRL19,DBLP:conf/cvpr/DaiWGJ21}. 
$\kappa$-GCN \cite{DBLP:conf/icml/BachmannBG20} extends GCN to  constant curvature manifolds.
In ultrahyperbolic or quotient spaces,  \cite{DBLP:conf/nips/XiongZPP0S22,DBLP:conf/nips/Law21} study node embedding in the time-space coordinates of some curvature radius.
\cite{DBLP:conf/icml/LopezPT0W21} explores the symmetric matrix manifold for graphs.
\cite{DBLP:conf/nips/ZhuP00C020,DBLP:conf/www/YangCPLYX22} represent graphs with the dual spaces of Euclidean and hyperbolic ones.
\cite{sun21aaai} considers graph dynamics in hyperbolic space, while \cite{sun23aaai,sun22cikm} model graph sequences in Riemannian manifolds.
\cite{sun23jmlc} studies bipartite graph in evolving  manifolds.
 \cite{DBLP:conf/iclr/GuSGR19,sun22aaai}  study graph representation learning in the product space.
However, they do not model Ricci curvatures with the product space for structure learning.
Also, we notice that Ricci curvature is recently introduced for graph clustering \cite{sun23ijcai}, graph rewriting \cite{DBLP:conf/iclr/ToppingGC0B22} and edge dropping \cite{DBLP:conf/www/Liu0P00C023}, 
 but none of them learn a graph structure in absence of a differentiable Ricci curvature formulation.
Distinguishing from the existing studies, we introduce a latent Riemannian space with gyrovector map for GSL.


\vspace{-0.1in}
\section{Conclusion}
\vspace{-0.05in}
In this paper, we pose a new problem of self-supervised graph structure-feature co-refinement, 
and propose \textbf{DeepRicci} from a fundamentally different perspective of Riemannian geometry, 
alleviating over-squashing with theoretical guarantees.
Specifically, we introduce a novel latent Riemannian space with gyrovector  feature mapping, 
where we refine structure by backward Ricci flow with differentiable Ricci curvature, and simultaneously refine feature by geometric contrastive learning among different Riemannian factors.
Extensive experiments on public datasets show the superiority of  DeepRicci.

\vspace{-0.05in}

\vspace{-0.03in}
\section*{Acknowledgment}
\vspace{-0.05in}
The authors were supported in part by National Natural Science Foundation of China under grant 62202164,  U21B2027, 61972186, and 62266027, the Fundamental Research Funds for the Central Universities 2022MS018, Natural Science Foundation of Beijing Municipality through grant 4222030, Yunnan Provincial Major Science and Technology Special Plan Projects through grants 202302AD080003, 202202AD080003 and 202303AP140008, General Projects of  Basic Research in Yunnan Province through grant 202301AT070471, and 202301AS070047.
Philip S. Yu is supported in part by NSF under grant III-2106758. 
Corresponding Author: Zhengtao Yu.

\vspace{-0.03in}
\scriptsize
\bibliographystyle{IEEEtran}
\bibliography{icdm23}

\end{document}